\documentclass[letterpaper]{article} % DO NOT CHANGE THIS
\usepackage{aaai25}  % DO NOT CHANGE THIS
\usepackage{times}  % DO NOT CHANGE THIS
\usepackage{helvet}  % DO NOT CHANGE THIS
\usepackage{courier}  % DO NOT CHANGE THIS
\usepackage[hyphens]{url}  % DO NOT CHANGE THIS
\usepackage{graphicx} % DO NOT CHANGE THIS
\usepackage{natbib}  % DO NOT CHANGE THIS AND DO NOT ADD ANY OPTIONS TO IT
\usepackage{caption} % DO NOT CHANGE THIS AND DO NOT ADD ANY OPTIONS TO IT
\usepackage{amsmath}
\usepackage{algorithm}
\usepackage{algorithmic}
\usepackage{xcolor}
\usepackage{pifont}

\usepackage{newfloat}
\usepackage{listings}
\DeclareCaptionStyle{ruled}{labelfont=normalfont,labelsep=colon,strut=off} % DO NOT CHANGE THIS
\floatstyle{ruled}
\newfloat{listing}{tb}{lst}{}
\floatname{listing}{Listing}
\title{Finding Closure:\\ A Closer Look at the Gestalt Law of Closure in Convolutional Neural Networks}
\author{
    Yuyan Zhang\textsuperscript{\rm 1}\equalcontrib,
    Derya Soydaner\textsuperscript{\rm 2}\equalcontrib \textsuperscript{†}, 
    Lisa Ko\ss mann\textsuperscript{\rm 2},
    Fatemeh Behrad\textsuperscript{\rm 2},
    Johan Wagemans\textsuperscript{\rm 2} %\thanks{With help from the AAAI Publications Committee.}
}
\affiliations{
    \textsuperscript{\rm 1}Department of Computer Science, University of Leuven (KU Leuven), Leuven, Belgium\\
    \textsuperscript{\rm 2}Department of Brain \& Cognition, University of Leuven (KU Leuven), Leuven, Belgium\\
    \textsuperscript{†} Corresponding author: derya.soydaner@kuleuven.be
}

\usepackage{bibentry}
\begin{document}

\maketitle

\begin{abstract}
The human brain has an inherent ability to fill in gaps to perceive figures as complete wholes, even when parts are missing or fragmented. This phenomenon is known as Closure in psychology, one of the Gestalt laws of perceptual organization, explaining how the human brain interprets visual stimuli. Given the importance of Closure for human object recognition, we investigate whether neural networks rely on a similar mechanism. Exploring this crucial human visual skill in neural networks has the potential to highlight their comparability to humans. Recent studies have examined the Closure effect in neural networks. However, they typically focus on a limited selection of Convolutional Neural Networks (CNNs) and have not reached a consensus on their capability to perform Closure. To address these gaps, we present a systematic framework for investigating the Closure principle in neural networks. We introduce well-curated datasets designed to test for Closure effects, including both modal and amodal completion. We then conduct experiments on various CNNs employing different measurements. Our comprehensive analysis reveals that VGG16 and DenseNet-121 exhibit the Closure effect, while other CNNs show variable results. We interpret these findings by blending insights from psychology and neural network research, offering a unique perspective that enhances transparency in understanding neural networks. Our code and dataset will be made available on GitHub.
\end{abstract}


\section{Introduction} 

Tasks that come naturally to humans are still challenging for machines, despite the increasing integration of AI-based technologies across various domains. %such as image generation \cite{betker2023improving} and language modelling \cite{achiam2023gpt}. 
The ease with which neural networks can be confused, such as misclassifying an image of a cat textured as an elephant \cite{geirhos2018imagenet}, serves as an amusing yet stark reminder of their limitations compared to the human brain’s capabilities. While we can easily recognize faces, even a slight perturbation in an image can deceive a neural network \cite{szegedy2013intriguing, goodfellow2014explaining}, illustrating the significant gap that still exists in mimicking human visual perception.

Building on this, we tackle the Gestalt laws of perceptual organization, which have shaped our understanding of how visual information is grouped and processed \cite{wertheimer1923laws, palmer2002, wagemans2012}. Gestalt psychology has discovered principles of how things are grouped, explaining how we perceive an overall essence in visual stimuli before grasping the details. One of these principles is Closure, which states that individuals perceive incomplete shapes as complete, forming a coherent whole. Recent studies on CNNs have varied: some indicate that they exhibit Closure within certain thresholds and limitations %\cite{amanatiadis2018understanding, ehrensperger2019evaluating, kim2021}, 
while others do not. %\cite{baker2018deep, Zhang2024}. 
This lack of consensus and the diverse findings have prompted our research to further investigate and settle the debate on whether CNNs exhibit this Gestalt principle.

While both humans and CNNs can perform object recognition, humans naturally exhibit Closure. This study explores whether CNNs can also demonstrate this perceptual skill. By examining how neural network representations align with human perceptual processes, we enhance our understanding and inform the design of human-compatible AI systems \cite{sucholutskyshould}. Although CNNs operate differently from the human brain, which relies on recurrent interactions in cortical layers, understanding whether CNNs can approximate this perceptual task—despite their distinct architectures—could %deepen our understanding of AI systems, 
bridge the gap between human cognitive processes and machine learning techniques. Our main contributions are as follows: 
\begin{itemize}
   \item We present well-curated datasets designed to test Closure effects, with carefully controlled conditions to evaluate both modal and amodal completion in neural networks.
   \item We design psychology-based experiments to assess whether CNNs exhibit the Closure effect, addressing limitations and resolving the lack of consensus in literature.
   \item We conduct a detailed analysis of various CNNs, expanding the range of models examined in previous studies.
   \item Our research sheds light on the Gestalt law of Closure in CNNs, contributing to our understanding of their ability to mimic human visual perception. 
\end{itemize}

\section{Related Work}

An early study \cite{amanatiadis2018understanding} investigated six core Gestalt laws by training AlexNet \cite{krizhevsky2017imagenet} and Inception V1 (GoogLeNet) \cite{szegedy2015going} on the MNIST \cite{lecun1998gradient} and ImageNet \cite{deng2009imagenet} datasets. Their measurement for assessing Closure is based on the occlusion percentage. Their findings indicate that the Closure principle is effective in CNNs up to approximately 30\% occlusion, beyond which the models' performance decreases. Building on this accuracy-based method, a recent study expanded the investigation by training a broader range of CNNs on complete polygons and testing their performance on incomplete ones \cite{Zhang2024}. This study suggests that CNNs do not display the Closure effect when assessed with that particular measurement and dataset. %However, further investigation is necessary to draw conclusions.    

Another study fine-tuned AlexNet to classify wireframes and Kanizsa squares as `fat' and `thin' in a  shape discrimination task \cite{baker2018deep}. They concluded that neural networks do not perceive illusory contours. Subsequent studies, such as \cite{biscione2023mixed}, employed a similarity-based method and examined more models, including ResNet-152 \cite{7780459} and DenseNet-201 \cite{huang2017densely}. They found mixed evidence of perceptual grouping. Another study analyzed the classification performance of AlexNet and Inception V1 on Kanizsa triangles (we discuss Kanizsa triangles in the next section) and modified triangles with removed edge sections, reporting evidence of Closure \cite{ehrensperger2019evaluating}. Further research examined Inception and a simple CNN, both trained on natural images, and found that they exhibit Closure on synthetic edge fragments \cite{kim2021}. They tested these CNNs on incomplete triangles, finding they were more likely recognized as similar to complete triangles rather than disordered fragments. While not directly aligned with our research focus, a noteworthy study tested a latent noise segmentation model on various datasets reflecting Gestalt laws, including Closure \cite{lonnqvist2023latent}.

\section{Methodology}

\subsection{Problem Definition}  
\begin{figure}[ht]
    \centering    \includegraphics[width=0.4\textwidth]{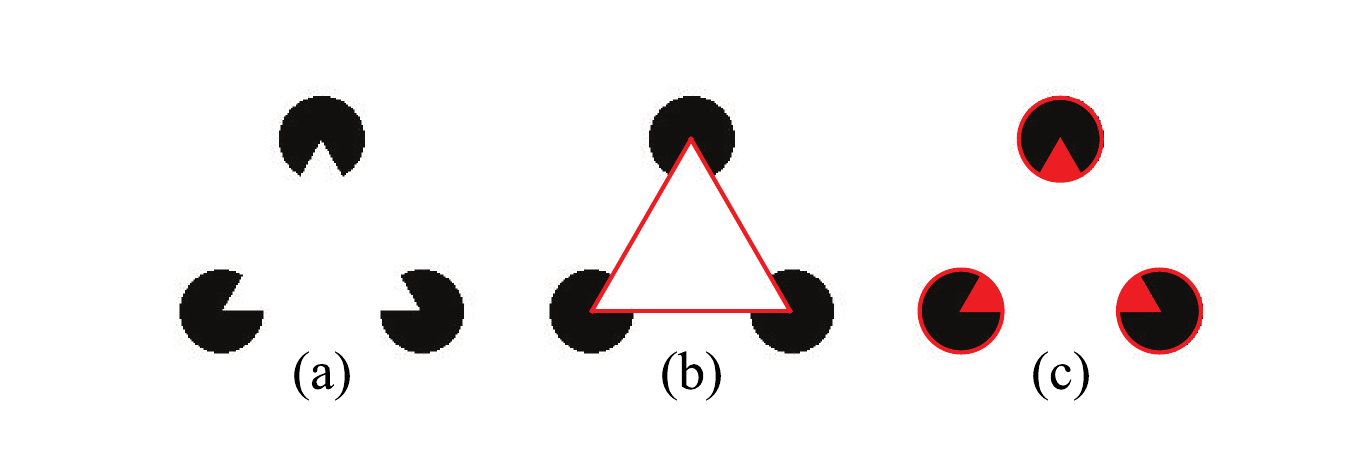}
    \caption{\emph{(a)} An example of a Kanizsa triangle. The image consists of three black fragments with a white triangle perceived in front them. \emph{(b)} Modal completion and \emph{(c)} Amodal completion in the Kanizsa triangle.}
    \label{fig:kanizsa}
\end{figure}

%\lisa{Closure Part is rather long now, but we can discuss which parts to cut together. Did not follow the thesis fully here was focused on telling a story}
The Gestalt principle of Closure, the tendency for the human visual system to prefer closed shapes over disjointed elements, is an integral part of human visual perception \cite{wagemans2012}. According to the Closure principle, the visual system completes contours to create closed shapes, through processes referred to as contour integration (distinct elements are integrated into a contour forming a shape) or contour completion (integrating smooth contours that are separated due to occlusion or camouflage) \cite{wagemans2012}. There are two forms of contour completion: amodal and modal completion \cite{Michotte1964, Wagemans2006}. During amodal completion, the object is completed behind an occluder. This has often been studied using line drawings and observers report the strong intuition that the object continues behind the occluder \cite{Wagemans2006}. Modal completion includes so-called illusory contours and surfaces, which are triggered by specific stimulus characteristics. A prominent example of modal completion is the Kanizsa Triangle \cite{Kanizsa1955} shown in Figure \ref{fig:kanizsa}. Here, the specific properties of the “pac-man” shaped circles, particularly their missing parts, give rise to a white triangle superimposed on three black circles. Note, however, while distinctions between the two processes can be made, they often coincide. In the Kanizsa triangle, for example, amodal completion is also present, in the form of the circles being completed behind the triangle \cite{Wagemans2006}. 

Line drawings and Kanizsa triangles may appear to be specific instances that bear little importance in the complex visual reality of everyday life. However, the processes the study of their perception reveals, are highly important for understanding how the human visual system organizes the -sometimes ambiguous- input from the optic apparatus into objects \cite{wagemans2012} and enables us to understand whole scenes in the matter of milliseconds \cite{Potter2014}. Therefore, investigating this essential human visual `skill' in neural networks, can not only shed light on their comparability with humans but also could its integration potentially increase their detection and classification abilities \cite{Zhang2024}.

\subsection{Convolutional Neural Networks} 

CNNs have revolutionized computer vision by learning spatial features directly from images. Pioneering models such as AlexNet \cite{krizhevsky2017imagenet} and VGG16 \cite{simonyan2014very} have achieved high accuracy in image recognition tasks. However, their use of numerous small filters has resulted in high computational complexity. Later advances addressed this issue: ResNet \cite{7780459}, SqueezeNet \cite{iandola2016squeezenet}, and DenseNet \cite{huang2017densely} employ residual connections, efficient filters, and dense connectivity, respectively, to achieve comparable or superior accuracy while reducing computational costs. Other progress include Inception networks \cite{szegedy2015going} and ShuffleNet \cite{zhang2018shufflenet}. Inception networks use filters of varying sizes within each layer to capture a wider range of spatial features. ShuffleNet enhances information flow through channel shuffling, significantly improving performance. Additionally, EfficientNet \cite{tan2019efficientnet} uses compound scaling to optimize network depth, width, and resolution based on resource constraints. MobileNet \cite{howard2017mobilenets}, featuring depthwise separable convolutions, is particularly well-suited for mobile devices due to its efficient processing.

In our experiments, we utilized pre-trained CNNs, freezing all their layers. We refrained from training them on completion tasks, as training on natural images resembles the development of visual perception in humans. Humans exhibit Gestalt laws without explicit training. Our research explores whether these neural networks can similarly exhibit such inherent capabilities. We obtain our results based on the output of the last convolutional layer in VGG16, SqueezeNet V1.1, ShuffleNet V2, and MobileNet V3; the final MBCConv6 layer for EfficientNet B0; Mixed 7c layer for Inception V3; the last fully-connected layer for AlexNet; the final average pooling layer for ResNet-50; and the final dense block for DenseNet-121.

%We did not explicitly train them on completion tasks because training them on natural images resembles the development of visual perception in humans. Humans can exhibit the effects of Gestalt rules without explicit training, and we are interested in whether these neural networks can do the same.

\section{Experiments} 

\subsection{Experiment 1: The Similarity-Based Method} 
 
Our first experiment builds upon the work of Kim \emph{et al.} \cite{kim2021}, who demonstrated that CNNs trained on natural images do perform Closure. We begin by replicating their study, generating the same dataset with two necessary modifications and extending the methodology to a broader range of CNNs. This part focuses on Triangle Segment Completion, using displays of edge fragments to demonstrate that a CNN trained for classification exhibits Closure \cite{kim2021}. This extended replication enables us to determine whether Closure can be observed in a broader range of CNNs beyond just the Inception network and sets a baseline for our subsequent analysis. We then adapt the experiment to include Kanizsa triangles—a condition not previously explored in the referenced study. 

\subsubsection{Triangle Segment Completion}
%The purpose of this experiment is to replicate the findings of Kim et al. (2021), who used displays of edge fragments to show that a CNN trained for classification exhibits Closure \cite{kim2021}. We will use the replication of their work as a baseline for our experiment.

%\begin{figure}[ht]
%\centering
%\includegraphics[width=0.9\columnwidth]{figures/dataset2.pdf} % Reduce the figure size so that it is slightly narrower than the column. Don't use precise values for figure width.This setup will avoid overfull boxes.
%\caption{Examples of 3 groups of images used in the first experiment, which shared the same set of parameters as those used by Kim et al. (2021).}
%\label{dataset2}
%\end{figure}

\subsubsection{Dataset.} 
\begin{figure}[ht]
\centering
\includegraphics[width=0.9\columnwidth]{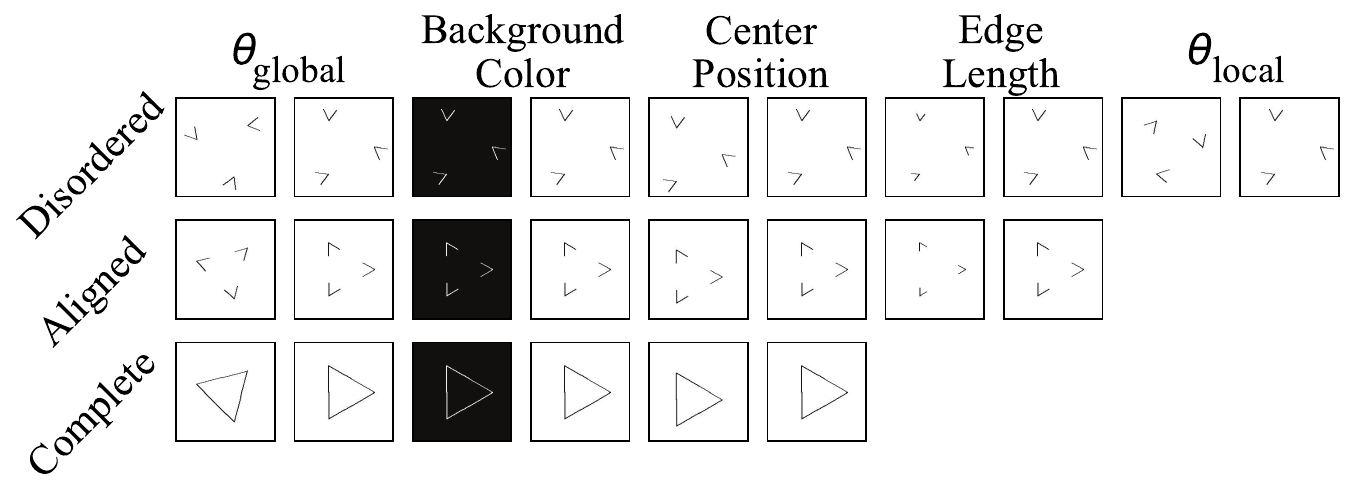} % Reduce the figure size so that it is slightly narrower than the column. Don't use precise values for figure width.This setup will avoid overfull boxes.
\caption{Examples of the three groups of images used in the first experiment, with parameters matching those specified by Kim \emph{et al.} 
%utilizing the same set of parameters as those specified by Kim \emph{et al.} 
\cite{kim2021}.}
\label{dataset2}
\end{figure}

The dataset used by Kim \emph{et al.} \cite{kim2021} contains 992 different images, which are categorized into three groups: %$\colon$ 
complete triangles, aligned triangle fragments, and disordered triangle fragments. The triangles and fragments differ in global orientations ($\theta_{global}$), background colors, center positions, edge length (defined as the length of the visible edge of each fragment in the image; 3, 8, 13, 18, 24, or 29 pixels, respectively) and local orientations ($\theta_{local}$) (see Figure~\ref{dataset2}). $\theta_{global}$ is the degree to which the triangle rotates around its center in the plane, with values from   0\textdegree, to 105\textdegree in steps of  15\textdegree. ($\theta_{local}$), describing the degree to which the triangle fragment rotates around the corresponding vertex, can be 72\textdegree, 144\textdegree, 216\textdegree, or 288\textdegree. All fragments in an image share the same $\theta_{local}$ and have a size of $150 \times 150$. The distance between any two vertices in any image is 116 pixels. Overall, there are 32 complete, 192 aligned, and 768 disordered triangles.
We replicate their dataset with two notable modifications: (1) the size of each image is set to $300 \times 300$. (2) the possible positions of the center of a triangle were changed to (150, 150) and (134, 134).   
%\begin{itemize}
%    \item the size of each image is set to $300 \times 300$.
%    \item the possible positions of the center of a triangle were changed to (150, 150) and (134, 134).
%\end{itemize} 
This is because in some conditions (e.g., when the edge length is 29 pixels, $\theta_{global}$ = 0\textdegree, and $\theta_{local}$ = 144\textdegree), a disordered triangle fragment would go over the border of the canvas and the resulting image would not be able to show the whole fragment.

\subsubsection{Measurement.} We adopt the measurement for assessing the Closure effect as implemented by Kim \emph{et al.} \cite{kim2021} to ensure a fair comparison across different CNN models. This measurement is shown below, where \emph{Similarity} represents cosine similarity:   

\begin{multline}
    \text{Closure Measure} = \text{Similarity}(\text{aligned}, \text{complete}) \\
    - \text{Similarity}(\text{disordered}, \text{complete}) \label{eq: Closure Measure}
\end{multline}

\begin{equation}
    Similarity(\mathbf{x}, \mathbf{y}) = \frac{f(\mathbf{x}) f(\mathbf{y})^T}{|f(\mathbf{x})| | f(\mathbf{y})|}
\end{equation}
% $$\text{Closure Measure} = 
% \text{Similarity}(\text{aligned}, \text{complete}) - \text{Similarity}(\text{disordered}, \text{complete})$$
%where \text{Similarity} represents cosine similarity: 
%$$Similarity(\mathbf{x}, \mathbf{y}) = \frac{f(\mathbf{x}) f(\mathbf{y})^T}{|f(\mathbf{x})| | f(\mathbf{y})|}$$
where $ f(\mathbf{x})$ is the output vector of a layer in the model. 

If the model exhibits Closure, we would expect the similarity value between a pair of aligned and complete triangles to be greater than that between a pair of disordered and complete triangles, resulting in a relatively large difference. 

\subsubsection{Results and Discussion.} 

\begin{figure}
    \centering    \includegraphics[width=0.85\linewidth]{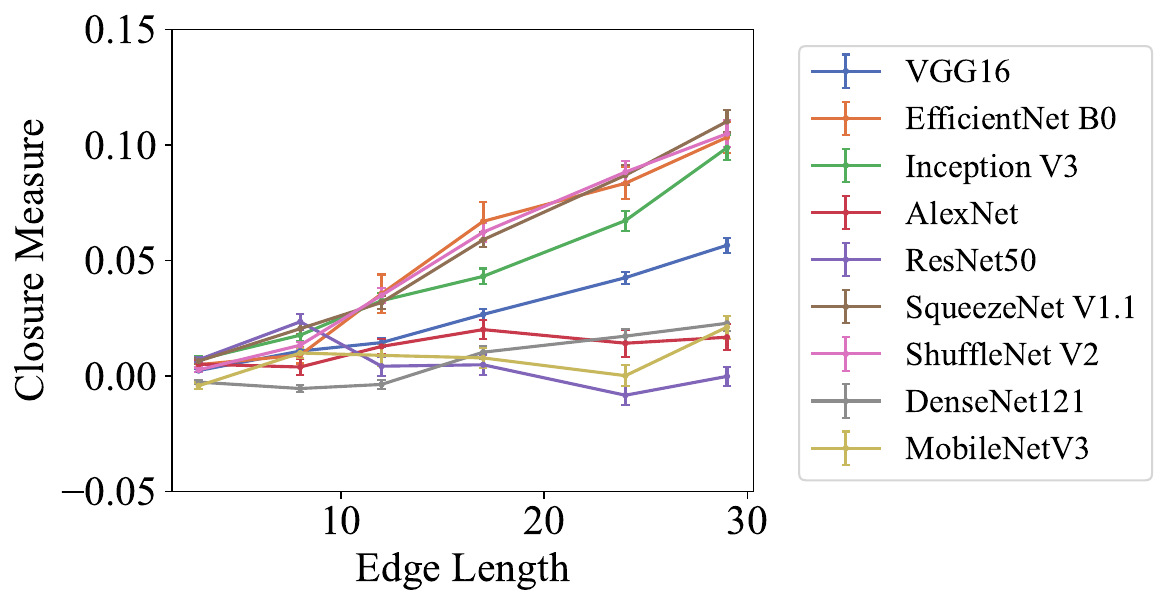}
    \caption{Closure measure plotted against edge length (in pixels) for Experiment 1, `Triangle Segment Completion'. The Closure measure ranges from -1 to 1, with larger values indicating a stronger Closure effect. %(b) illustrates the locality measure versus the edge length (in pixel). Locality measure also ranges from -1 to 1 and a larger value suggests a stronger focus on local features instead of global features. In both features, each point represents the mean value for images at each edge length and the error bar implies the standard error. Each color suggests the results from a different model.
    }
    \label{fig:results2}
\end{figure}

The average values of the Closure measure for AlexNet, ResNet-50, DenseNet-121, and MobileNet V3 are around 0 for all edge lengths (Figure \ref{fig:results2}). However, the Closure measure increases with edge length increases for VGG16, EfficientNet B0, Inception V3, SqueezeNet V1.1, and ShuffleNet V2. Most of their Closure values exceed 0 when the edge length is 13 pixels or more. 

To further explore whether there is a significant increasing trend in the Closure measure, we conduct a general multivariate linear regression analysis for each model. The predicted value is the Closure measure, and the predictors are the edge length, the $\theta_{global}$, the background colour, the centre position, and the $\theta_{local}$. Among these variables, the Closure measure and the edge length are treated as continuous, while all other variables are nominal. The results show that although all regression models are significant ($p < .001$), only those for VGG16, EfficientNet B0, Inception V3, SqueezeNet V1.1, and ShuffleNet V2 have at least moderate effect sizes (adjusted $R^2 > .40$). For these models, the edge length can significantly predict the Closure measure when other variables are controlled ($b > .0030$, $p < .001$). The regression models for AlexNet and DenseNet-121 have small effect sizes (adjusted $R^2 < .30$) and the predicting effect of the edge length is significant in the two regression models ($b = .0003$, $p < .001$ for AlexNet, and $b = .0006$, $p < .001$ for DenseNet-121).

To summarize, our first experiment not only replicates the results of Kim \emph{et al.} \cite{kim2021} on Inception V3 but also demonstrates the Closure effect in VGG16, EfficientNet B0, SqueezeNet V1.1, ShuffleNet V2, AlexNet, and DenseNet-121, although the effects of the last two models
might be small. On the other hand, ResNet-50, and MobileNet V3 do not show the Closure effect. This difference in the results could be due to the structures of the models. However, the type of Closure or the measurement may also play a role. Further experiments are conducted to explore these possibilities, discussed in the following sections. 

\subsubsection{Kanizsa Triangles}

% Following concerns that Kim et al. (2021) might only measure simple pixel-wise comparisons, this section tests the models' ability to generalize the Closure effect to Kanizsa triangles.

\subsubsection{Dataset.} 
\begin{figure}[ht]
\centering
\includegraphics[width=0.9\columnwidth]{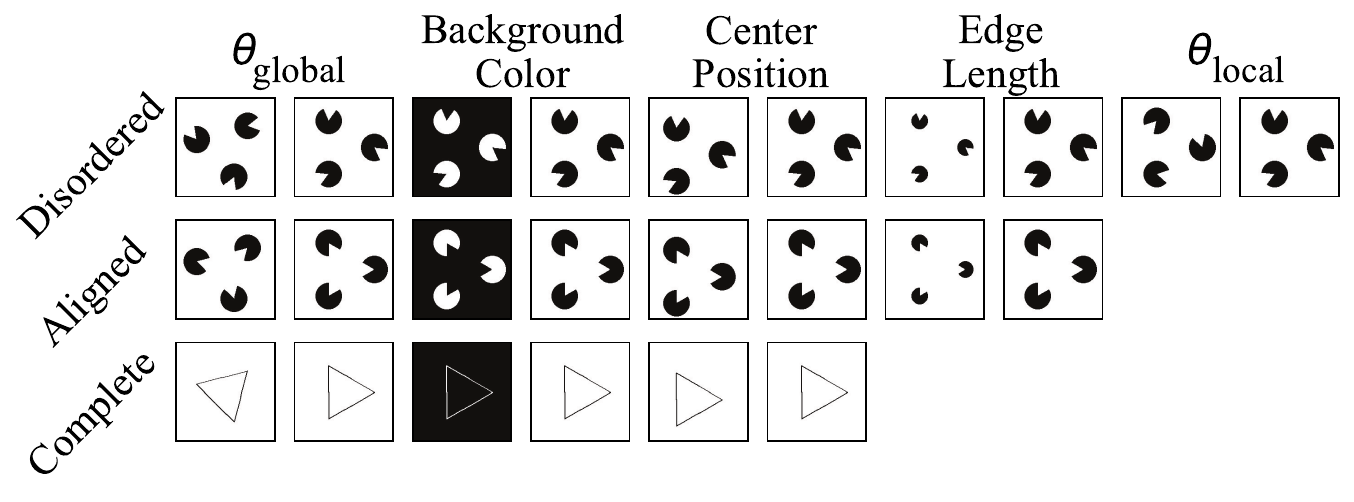} % Reduce the figure size so that it is slightly narrower than the column. Don't use precise values for figure width.This setup will avoid overfull boxes.
\caption{Examples of the three groups of images featuring Kanizsa triangles used in our experiment.}
\label{dataset3}
\end{figure}

The dataset for this experiment has the same size and settings as the previous one except that it includes complete triangles, valid Kanizsa triangles (aligned triangle fragments), and invalid Kanizsa triangles (disordered triangle fragments), as illustrated in Figure \ref{dataset3}. While all other parameters remain consistent with the previous experiment, the valid Kanizsa triangles in this dataset are composed of three incomplete disks, rather than triangle fragments. 

\subsubsection{Measurement.}
The Closure measure is defined as the difference in similarity between a valid Kanizsa triangle and a complete triangle, compared to an invalid Kanizsa triangle and a complete triangle. A large positive value suggests a strong Closure effect, while a value close to 0 indicates no Closure effect. The rationale is that if a model exhibits the Closure effect, it would perceive an illusory contour of a triangle in the valid condition, leading to more similar representations of a valid Kanizsa triangle and a complete triangle. We use the same models and layers as in the Triangle Segment Completion part of this experiment.

\subsubsection{Results and Discussion.}
\begin{figure}
    \centering    \includegraphics[width=0.85\linewidth]{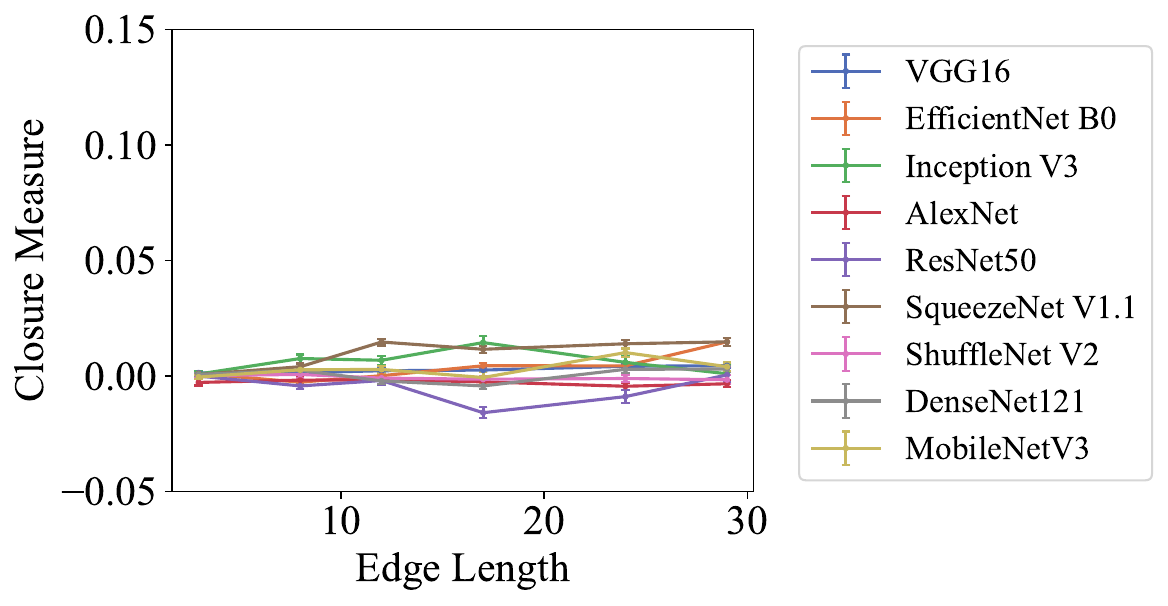}
    \caption{Closure measure plotted against edge length (in pixels) for Experiment 1, `Kanizsa Triangles'. The Closure measure ranges from -1 to 1, with larger values indicating a stronger Closure effect. %(b) illustrates the locality measure versus the edge length (in pixel) for the Kanizsa triangle dataset. Locality measure also ranges from -1 to 1 and a larger value suggests a stronger focus on local features instead of global features. In both features, each point represents the mean value for images at each edge length and the error bar implies the standard error. Each color suggests the results from a different model and the assignment of colors is the same as that in Figure \ref{fig:results2}.
    }
    \label{fig:results3}
\end{figure}

Results (Figure \ref{fig:results3}) show a negligible Closure effect across all models and edge lengths.
Multivariate linear regression analyses revealed significant but weak effect sizes (adjusted $R^2 < .16$) on the Closure measure for Inception V3, AlexNet, ResNet-50, ShuffleNet V2, DenseNet-121, and MobileNet V3. Among these, the predictive effect of edge length on the Closure measure is significant for ShuffleNet V2, DenseNet121, and MobileNet V3, but their coefficients are much smaller than those obtained in the previous experiment ($b < .0004$, $p < .001$). The regression models for VGG16, EfficientNet B0, and SqueezeNet V1.1 are significant ($p < .001$) and have medium effect sizes (adjusted $R^2 > .30$). Additionally, edge length significantly predicts the Closure measure in these models ($p < .001$), although the coefficients are smaller than those observed for each model in the Triangle Segment Completion experiment ($b = .0006$, $.001$, and $.0009$, respectively).  

These results suggest that Inception V3, AlexNet, and ResNet-50 cannot perceive the illusory contour in Kanizsa triangles under the current similarity-based measurement. VGG16, EfficientNet B0, SqueezeNet V1.1, ShuffleNet V2, DenseNet-121, and MobileNet V3 exhibit Closure effects, but these effects are much smaller than those observed for incomplete triangles in Triangle Segment Completion. 

However, the possibility remains that the current method is not sensitive enough to detect the capability of the models to perceive illusory contours. It has to be noted, for the Kanizsa triangles, we cannot rule out a section of overlap given the larger space the pacmen occupy. We tried to address this by randomly adjusting $\theta_{global}$ and center positions. Regardless, some overlap persists, potentially diluting the results. Given that the overlapping area for Kanizsa triangles is much smaller than the non-overlapping area, this measure is still valid, though it should be interpreted with caution.
%albeit having to be interpreted more cautiously. 
Therefore, we conducted further experiments to investigate whether the models could perceive illusory contours or perform line completion under different measurements. 
%While it can not be ruled out that the similarity-based measurement we utilize is limited in its sensitivity, 
Although the sensitivity of the similarity-based measurement may be limited, enhancing the dataset with Kanizsa triangles is an important contribution and makes a stronger case for models exhibiting the Closure effect. Kanizsa triangles include modal and amodal completion and are therefore a highly complex stimulus considering Closure.

\subsection{Experiment 2: The Configural Effects (CE)-Based Method} 
Our second experiment uses the concept of configural effects (CE; \cite{pomerantz1977perception}), which suggests that the perception of a stimulus is influenced by its overall configuration rather than only its individual components. %This concept is a key aspect of Gestalt psychology, which highlights the key principle that `the whole is greater than the sum of its parts' \cite{koffka1935}.

Recently revisited by \cite{biscione2023mixed}, we have adapted this concept to assess the Closure effect, utilizing a novel dataset in our analysis. Each set in this dataset is composed of two pairs of images: a base and a composite pair. Our hypothesis posits that if a model exhibits Closure, it differentiates composite pairs more easily than base pairs. Consequently, the dissimilarity value of a base pair should be smaller than a composite pair. This would result in a positive difference between the dissimilarity values of the composite pair and the base pair, thus confirming the presence of the Closure effect. 

We test this hypothesis on the same CNNs and layers used in the previous experiment. Our approach focuses on analyzing the internal representations of images within these models, rather than directly using the predicted classes. To do this, we compare the activation functions of these CNN layers across different sets of images. %Our dataset, the measurement for this experiment, and the results are detailed below.   

\subsubsection{Dataset.}
\begin{figure}[ht]
\centering
\includegraphics[width=0.8\columnwidth]{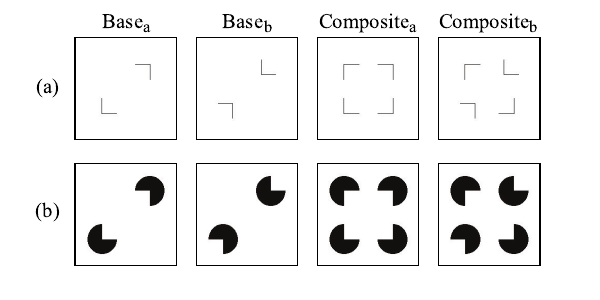} % Reduce the figure size so that it is slightly narrower than the column. Don't use precise values for figure width.This setup will avoid overfull boxes.
\caption{Examples of images used in the second experiment. \emph{(a)} the line segments condition. \emph{(b)} the Kanizsa squares condition. The left two images in each condition are the base pair ($\text{base}_a$ and $\text{base}_b$) and the right two are the composite pair ($\text{composite}_a$ and $\text{composite}_b$).}
\label{dataset4}
\end{figure}

Our dataset is illustrated in Figure~\ref{dataset4}, where we test the Closure effect under two conditions: the line segments condition and the Kanizsa squares condition. For each condition, one set of images consists of two pairs: the base pair and the composite pair. The base pair contains two different images ($\text{base}_a$ and $\text{base}_b$) and either of the two shapes in one image is different from that at the corresponding location in the other image. When two additional components are added to each image in the base pair (the added components are the same for both images in the base pair), they form the composite pair ($\text{composite}_a$ and $\text{composite}_b$). 

Each component in the line segments condition is a fragment of a square, while in the Kanizsa squares condition, each component is an incomplete disk. For each composite image in the line segments condition, the four components are located so that the distance between the centroids of any two neighboring components is the same. The base images contain opposing components in the corresponding composite images. By keeping the distance between centroids instead of vertices to be the same, the distances between the two components in the base pair images would be perceived as the same, and the composite pair images would be perceived as ``aligned'' by humans. In the line segments condition, the distance between any two neighboring vertices in $\text{composite}_a$ is 95 pixels. In the Kanizsa squares condition, the distance between the centers of neighboring incomplete disks is always 95 pixels. All other parameters are the same across the images in both conditions. 

In each condition, the sets of images differ from other sets in their $\theta_{global}$, edge lengths, background colors, and center positions. The $\theta_{global}$ of an image can take any of the following values: 0\textdegree, 11.25\textdegree, 22.5\textdegree, 33.75\textdegree, 45\textdegree, 56.25\textdegree, 67.5\textdegree, or 78.75\textdegree. Larger degrees are not used because the resulting image would be identical to one generated with a previous $\theta_{global}$. The edge length of an image can be one of the following values: 5, 10, 14, 19, 24, 29, 33, 38, or 43 pixels. In this case, the proportions of the edge length to the side length of the square correspond to 0.9, 0.8, 0.7, 0.6, 0.5, 0.4, 0.3, 0.2, and 0.1, respectively. The background color can be either white or black, and the center position can be either (150, 150) or (134, 134). The last two parameters match those in the first experiment in order to control for irrelevant variables. In total, there are 288 sets of images in each condition, consisting of 288 base pairs and 288 composite pairs. Thus, the number of images in each condition is 1152 and the total number of images used in the experiment is 2304. 

\subsubsection{Measurement.}
We employ the measurement inspired by the concept of CE \cite{pomerantz1977perception} and implemented by \cite{biscione2023mixed}, which is based on the Euclidean distance:

\begin{equation}
% \small 
 CE=\frac{D^l(composite_a, composite_b) - D^l(base_a, base_b)}{D^l(composite_a, composite_b) + D^l(base_a, base_b)}
\end{equation}
where $D^l(\mathbf{a}, \mathbf{b}) = || d^l(\mathbf{a}) - d^l(\mathbf{b}) ||$.

In human experiments, participants are asked to find the “odd” image among four presented images, consisting of three identical images and one different image. The pair of different images is either the base pair or the composite pair. If participants perform better on the composite pair than on the base pair (i.e., having shorter reaction times in the discrimination task for the composite pair than for the base pair), they are demonstrating Configural Superiority Effects (CSE) and utilizing Gestalt principles. 

\begin{figure*}
    \centering    
    \includegraphics[width=0.75\linewidth]{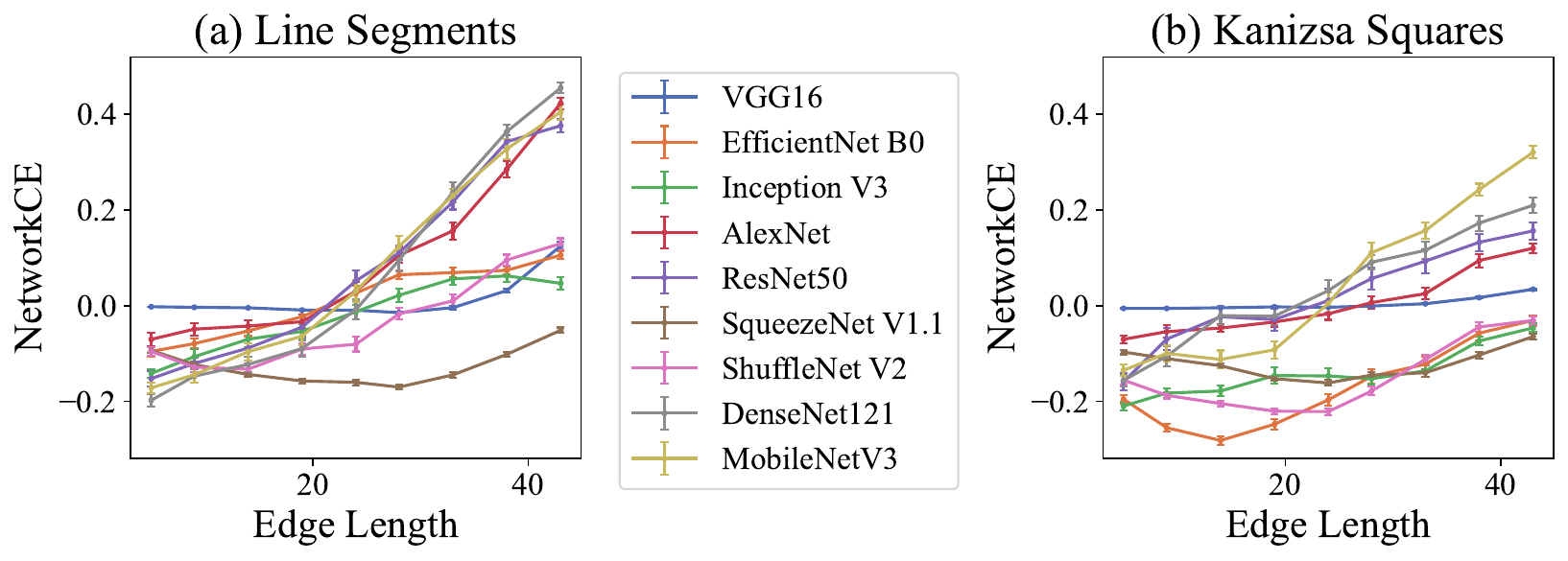}
    \caption{The line graphs with error bars showing network CE values under each edge length condition for all models. The error bars represent the standard errors of each model tested on images across edge lengths in each condition. A positive network CE value indicates the existence of the Closure effect in the corresponding condition, and a larger CE value implies a larger Closure effect. \emph{(a)} presents the results of the line segments condition and \emph{(b)} shows the results of the Kanizsa squares condition.}
    \label{fig:results4}
\end{figure*}

The key point of our experiment is that if the model exhibits the Closure effect, it should be easier for the model to differentiate the composite pair than the base pair. Consequently, the Euclidean distance of the base pair should be smaller than the composite pair. This results in a positive difference between the dissimilarity values of the base pair and the composite pair. Thus, a positive CE value indicates the presence of the Gestalt principle.

\subsubsection{Results and Discussion.}
Our results are depicted in Figure \ref{fig:results4}. In both conditions, most models show an increase in the CE values, which are positive when the edge length is large enough. Additionally, we conduct a one-sample t-test on the CE values for each model at each edge length in both the line segments and Kanizsa squares condition.

\emph{The line segments condition.} The average CE value for AlexNet is significantly larger than 0 when the edge length is 28 pixels or more ($p < .001$), reaching a mean of 0.042 at the edge length of 43 pixels. In ResNet-50, the average CE is significantly larger than 0 when the edge length is 24 pixels or more ($p = .024 < .05$ at 24 pixels, and $p < .001$ for larger edge lengths), with a maximal average CE of 0.376. DenseNet-121 presents a significant Closure effect ($p < .001$ for edge lengths of 28 pixels or more), with the largest average CE being 0.455. The average CE for MobileNet V3 is significantly positive when the edge length is 28 pixels or more, achieving a maximum value of 0.405. VGG16, EfficientNet B0, Inception V3, and ShuffleNet V2 show significantly positive average CE values when the edge length is sufficiently large, although the values are not as high as those in AlexNet, ResNet-50, DenseNet-121, and MobileNet V3. SqueezeNet V1.1 exhibits significantly negative CE values ($p <.001$), indicating that it does not demonstrate the Closure effect.

%because they represent the composite images as more different than the base images. 
%According to Biscione and Bowers (2023), if a model employs the Gestalt principle when processing one of the two images in the composite pair, the representations of the composite images are more differentiated than those of the base images. Assuming this hypothesis to be in accordance with the mechanism underlying the neural networks, the conclusion could be drawn that the models except for SqueezeNet V1.1 present the Closure effect when the removed proportion of a side in the square is not too much. 

The results above indicate that all models tested in the experiment, except for SqueezeNet V1.1, exhibit the Closure effect in the line segments condition, provided that the removed portion of a side in the square is not too large. Among those employing the Closure principle in processing the images, AlexNet, ResNet-50, DenseNet-121, and MobileNet V3 exhibit larger average CE values compared to the other four models, indicating stronger Closure effects. These models may perform better at differentiating aligned and disordered square segments, demonstrating a stronger ability to utilize the Closure principle.  

When the edge length is small (i.e., shorter than 24 pixels), all models except for VGG16 have average CE values that are significantly lower than 0. These results suggest that for those models, distinguishing between the two composite images becomes more difficult than differentiating the base pair. This difficulty may arise because the added components in the composite pair increase the image processing cost, making it harder for the models to distinguish between the images in the composite pair. 

\begin{table*}[!ht]
\centering
\begin{tabular}{l|l|ll|ll}
Model & V1 &\multicolumn{2}{c|}{Similarity-based method} & \multicolumn{2}{c}{CE-based method} \\ \cline{3-6}
 &  & Incomplete triangles & Kanizsa triangles & Incomplete squares & Kanizsa squares \\ \hline
VGG16 & 0.538 & \textcolor{black}{\ding{52}} ($r \leq 0.9$) & \textcolor{black}{\ding{51}} ($r \leq 0.8$) & \textcolor{black}{\ding{52}} ($r \leq 0.2$) & \textcolor{black}{\ding{52}} ($r \leq 0.3$) \\
EfficientNet B0 & 0.492 & \textcolor{black}{\ding{52}} ($r \leq 0.8$) & \textcolor{black}{\ding{51}} ($r \leq 0.7$) & \textcolor{black}{\ding{52}} ($r \leq 0.4$) & \textcolor{black}{\ding{55}} \\
Inception V3 & 0.496 & \textcolor{black}{\ding{52}} ($r \leq 0.9$) & \textcolor{black}{\ding{55}} & \textcolor{black}{\ding{52}} ($r \leq 0.3$)& \textcolor{black}{\ding{55}} \\
AlexNet & 0.508 & \textcolor{black}{\ding{51}} ($r \leq 0.7$) & \textcolor{black}{\ding{55}} & \textcolor{black}{\ding{52}} ($r \leq 0.4$) & \textcolor{black}{\ding{52}} ($r \leq 0.3$) \\
ResNet-50 & 0.511 & \textcolor{black}{\ding{55}} & \textcolor{black}{\ding{55}} & \textcolor{black}{\ding{52}} ($r \leq 0.5$) & \textcolor{black}{\ding{52}} ($r \leq 0.4$) \\
SqueezeNet V1.1 & 0.158 & \textcolor{black}{\ding{52}} ($r \leq 0.9$) & \textcolor{black}{\ding{51}} ($r \leq 0.9$) & \textcolor{black}{\ding{55}} & \textcolor{black}{\ding{55}} \\
ShuffleNet V2 & 0.446 & \textcolor{black}{\ding{52}} ($r \leq 0.9$) & \textcolor{black}{\ding{51}} ($r \leq 0.4$) & \textcolor{black}{\ding{52}} ($r \leq 0.2$) & \textcolor{black}{\ding{55}} \\
DenseNet-121 & 0.497 & \textcolor{black}{\ding{51}} ($r \leq 0.7$) & \textcolor{black}{\ding{51}} ($r \leq 0.6$) & \textcolor{black}{\ding{52}} ($r \leq 0.4$) & \textcolor{black}{\ding{52}} ($r \leq 0.4$) \\
MobileNet V3 & 0.499 & \textcolor{black}{\ding{55}} & \textcolor{black}{\ding{51}} ($0.5 \leq p \leq 0.6$) & \textcolor{black}{\ding{52}} ($r \leq 0.4$) & \textcolor{black}{\ding{52}} ($r \leq 0.4$) \\
\end{tabular}

\caption{Summary of results indicating the presence of the Closure effect in each model across two experiments. We include brain scores on V1 (primary visual cortex) for general discussion. A checkmark indicates the presence of the Closure effect, with thicker checkmarks denoting a moderate to large effect size and thinner checkmarks indicating a small effect size. Crosses indicate no Closure effect has been found. $r$ refers to the removal percentage and $r \leq a$ suggests that the Closure effect only exists when the removal percentage is no larger than $a$.}
\label{table1}
\end{table*}

\emph{The Kanizsa squares condition.} The average CE value for MobileNet V3 is significantly larger than 0 when the edge length is no less than 28 pixels ($p < .001$1), with the highest average CE being 0.321. DenseNet-121 also shows significantly positive average CE values at edge lengths of 28 pixels or more ($p < .001$), with a maximum average CE of 0.210. ResNet-50, AlexNet, and VGG16 show significantly positive average CE values when the edge length is larger or equal to 28, 33, and 33 pixels ($p < .001$), respectively. The largest average CE values for the three models are 0.156, 0.120, and 0.034, respectively. 

These results indicate that for MobileNet V3, DenseNet-121, ResNet-50, AlexNet, and VGG16, the valid and invalid Kanizsa squares in the composite pair are represented as more different than the images in the base pair. As discussed above, a model employing the Gestalt principle might have more differentiated representations of an image that induces the Gestalt effect and an image that does not, compared to the representations of a base pair \cite{biscione2023mixed}. Thus, these models exhibit the Closure effect when the edge length is large enough. 

Other models (i.e., EfficientNet B0, Inception V3, SqueezeNet V1.1, and ShuffleNet V2) show significantly negative average CE values no matter how large the edge length grows. Models that exhibit the Closure effect also have negative CE values when the edge length is small. Thus, in these conditions, the models cannot facilitate the process of the Kanizsa square in the composite pair; rather, their processing of images might be impaired by the added information in the composite pair compared to the base pair.

\section{General Discussion}

We present an overview of our results from both experiments in Table \ref{table1}. Notably, VGG16 and DenseNet-121 consistently demonstrate the capability to utilize the Closure principle in both experiments. EfficientNet B0, SqueezeNet V1.1, and ShuffleNet V2 exhibit the Closure effect using the similarity-based method, but their results are not consistent with the CE-based measurement. Conversely, AlexNet, ResNet-50, and MobileNet V3 show robust Closure effects under CE-based measurements but not with the similarity-based method. Inception V3, however, displays the Closure effect only in the line segments conditions and not in the Kanizsa shapes conditions.

We observe that changing the dataset from line segments to Kanizsa shapes for the same measurement results in a decreased proportion of models exhibiting the Closure effect, accompanied by smaller effect sizes (i.e., the Closure measure does not increase significantly in the figure, and regression analysis gives a small coefficient for edge length). Using the similarity-based method, the Closure effect is detected in 7 out of 9 models with line segments, with only 2 models displaying relatively small effect sizes. In contrast, all 6 models showing the Closure effect with Kanizsa triangles exhibit small effect sizes. With the CE-based method, 8 out of 9 models show the Closure effect for line segment images, but this number drops to 5 for Kanizsa squares.

One possible explanation for these findings is that CNNs may possess stronger capabilities for amodal completion and weaker abilities for modal completion. This could explain why some models exhibit the Closure effect in the line segments condition but not in the Kanizsa shape condition, which requires both types of completion, and why the effect is stronger with line segments.  
%Another explanation may relate to the properties of Kanizsa shapes, which require both modal and amodal completion. 
Additionally, CNNs have only been trained to assign a single class to an image rather than categorizing it into multiple classes. This means that CNNs consider all image features when determining the class, including those of the black disks in Kanizsa images, which are irrelevant to the perception of illusory contours and may introduce noise. If this is the case, tests assessing modal completion in neural networks should be further developed to avoid the influence of amodal completion.

We also examined the Brain Score \cite{Schrimpf2018}, a widely-used metric that measures the alignment of neural networks with the human brain\footnote{https://www.brain-score.org/vision/}. The score ranges from 0 to 1, with 1 indicating the closest alignment. We found no clear relationship between the similarity-based method results and the Brain Scores. However, scores on the `neural’ aspect, especially on `V1' (primary visual cortex), are strongly related to the CE-based method results. Models with low `V1' scores are less likely to exhibit the Closure effect with the CE-based method. This suggests that models more aligned with V1 are more likely to show the Closure effect, supporting the idea that early stages of visual processing involve configural information \cite{Fox2017}.

Lastly, our study focused on CNNs for two main reasons. The first is to address gaps identified in previous research conducted on CNNs. Secondly, CNNs operate differently from other neural networks, such as Vision Transformers (ViTs) \cite{dosovitskiy2020image}, which incorporate attention mechanisms. Given the complex interrelationships between attention mechanisms and perceptual grouping in the human brain \cite{wu2023neural}, ViTs require separate consideration.

\appendix
\section{Conclusion}

We conducted a detailed analysis of the Closure effect in CNNs. Due to their feed-forward only architecture, CNNs might not seem an intuitive choice to investigate a mechanism derived from the human brain. We believe, however, that an even stronger case could be made for studying the architecturally more simple CNNs. The reasoning is as follows: if a `simple' CNN relies on mechanisms similar to perceptual grouping mechanisms in humans, without recurrent connections, this is potentially a very interesting find for the fields of AI, Neuroscience, and Psychology adding to the understanding of grouping mechanisms. If it does not, attempts to implement such mechanisms could potentially increase their performance and make them less vulnerable to adversarial attacks. The stimuli used in our experiments, traditionally utilized to assess Closure in humans, provide a natural bridge to exploring these phenomena in CNNs.
%The stimuli used in our experiments are what has been used to identify Closure in humans so using them for CNN is a natural segway. 
 
We evaluated current CNNs without modifications to assess their inherent ability to perform Closure, given their proficiency in image classification. We chose not to train CNNs to perform Closure. As it is an innate human mechanism that enables us -through contour completion and integration- to detect objects, CNNs with object detection capabilities should either already posses this mechanism or not. %In future work, it might be interesting to try and enhance CNNs by training them to perform Closure and see if their performance increases.
Our exploration highlights the significant challenges these models still face. By comparing CNN performance to human vision, we gain a deeper understanding of their limitations. In future work, training CNNs to perform Closure could provide insights into improving their performance.

\bibliography{aaai25}

\end{document}